\pdfoutput=1

\documentclass[11pt]{article}

\usepackage{EMNLP2022}

\usepackage{times}
\usepackage{latexsym}

\usepackage[T1]{fontenc}

\usepackage[utf8]{inputenc}

\usepackage{microtype}

\usepackage{inconsolata}

\usepackage{times}
\usepackage{latexsym}
\usepackage{multirow}
\definecolor{tp1}{HTML}{66CC00}
\definecolor{tp2}{HTML}{009900}
\definecolor{tp3}{HTML}{999900}
\definecolor{tp4}{HTML}{FF0000}
\definecolor{tp5}{HTML}{FF6666}
\usepackage[T1]{fontenc}

\usepackage[utf8]{inputenc}
\usepackage{microtype}
\usepackage{graphicx}
\usepackage{amsmath}
\usepackage{amssymb}
\usepackage{booktabs}
\usepackage{array}
\newcolumntype{L}[1]{>{\raggedright\let\newline\\\arraybackslash\hspace{0pt}}m{#1}}
\newcolumntype{C}[1]{>{\centering\let\newline\\\arraybackslash\hspace{0pt}}m{#1}}
\newcolumntype{R}[1]{>{\raggedleft\let\newline\\\arraybackslash\hspace{0pt}}m{#1}}
\usepackage{amsfonts}
\usepackage{xspace}
\usepackage{pifont}
\usepackage{float}
\usepackage{amsmath}
\usepackage{mathtools}
\usepackage{commath}
\usepackage{caption, subcaption}
\usepackage{breakcites}
\usepackage{xargs}                      
\usepackage{arydshln}
\usepackage{soul}
\usepackage{lipsum}  
\usepackage{booktabs}
\newcommand{\tabitem}{~~\llap{\textbullet}~~}
\usepackage{tablefootnote}

\usepackage[belowskip=-10pt,aboveskip=3pt]{caption}
\title{Hierarchical3D Adapters for Long Video-to-text Summarization}

\author{Pinelopi Papalampidi \quad \quad \quad \quad
  Mirella Lapata \\
  Institute for Language, Cognition and Computation \\
    School of Informatics, University of Edinburgh \\
  \url{p.papalampidi@sms.ed.ac.uk},~~\url{mlap@inf.ed.ac.uk}
}

\begin{document}
\maketitle
\begin{abstract}
  In this paper, we focus on video-to-text summarization and
  investigate how to best utilize multimodal information for
  summarizing long inputs (e.g., an hour-long TV show) into long
  outputs (e.g.,~a multi-sentence summary).  We extend
  SummScreen~\cite{chen2021summscreen}, a dialogue summarization
  dataset consisting of transcripts of TV episodes with reference
  summaries, and create a multimodal variant by collecting
  corresponding full-length videos.  We incorporate multimodal
  information into a pre-trained textual summarizer efficiently using
  adapter modules augmented with a hierarchical structure while tuning
  only~3.8\% of model parameters. Our experiments demonstrate that
  multimodal information offers superior performance over more
  memory-heavy and fully fine-tuned textual summarization methods.
\end{abstract}

\begin{table*}[t]
\small        
\centering
\begin{tabular}{@{}l@{~~~~}lll@{}L{28em}@{}}
\hline
 & \multicolumn{1}{c}{Modality}  & \multicolumn{1}{c}{Input} 
 & \multicolumn{1}{c}{Output} & \multicolumn{1}{c}{Datasets} \\ \hline 
\multirow{2}{*}{text-to-text} & text & short & short &
XSum~\cite{narayan2018don}, CNN-DailyMail~\cite{nallapati2016abstractive},
NYT \cite{durrett-etal-2016-learning}, Gigaword \cite{napoles-etal-2012-annotated} \\
& text & long & long & SamSum~\cite{gliwa2019samsum},
QMSum~\cite{zhong2021qmsum}, SummScreen~\cite{chen2021summscreen} \\
\hline \hline
\multirow{3}{*}{video-to-video} & vision & short & short & OVP~\cite{de2011vsumm}, YouTube~\cite{de2011vsumm}, SumMe~\cite{gygli2014creating} \\
& vision/text & short & short & TVSum~\cite{song2015tvsum} \\
& vision/text(/audio) & long & long & LoL~\cite{fu2017video},
TRIPOD+~\cite{papalampidi2020movie} \\ \hline \hline
video-to-text & vision & long & short & TACoS~\cite{rohrbach2014coherent}\footnotemark \\
& vision/text/audio & short & short & How2~\cite{sanabria2018how2} \\
 & vision/text/audio & long & long & SummScreen\textsuperscript{3D}  \\
\hline
\end{tabular}
\caption{Summarization datasets grouped based on the input/output modalities and input/output length.}
\label{tab:datasets_overview}
\end{table*}

\section{Introduction}

What happens in the very last episode of ``Friends''? Anyone who has
seen this episode can summarize its key moments: Ross confesses his
love for Rachel, they decide to resume their relationship, while
Monica and Chandler adopt twins and move to the suburbs.  TV viewers
can naturally perform this dialogue summarization task having access
to multiple modalities: they not only hear the actors speak but also
see their expressions, actions, and whereabouts on screen.

Despite recent advances in summarization
\cite{nallapati2016abstractive,see-etal-2017-get,liu-lapata-2019-text}
and increasing interest in different types of dialogue summarization,
e.g.,~from meeting transcripts \cite{gliwa2019samsum,zhong2021qmsum}
or screenplays \cite{chen2021summscreen}, the contribution of
modalities other than text remains relatively understudied. This is
not entirely surprising given the challenges associated with the
multimodal summarization task illustrated above (e.g.,~produce a
written summary of a TV episode). Firstly, the input is long, it
cannot fit into standard sequence-to-sequence architectures, and the
different modalities have to be somehow combined; secondly, the output
is also long, summaries consist of multiple sentences and rich
vocabulary; and thirdly, it involves complex inference over long-range
dependencies between events and characters and common sense
reasoning. At the same time, creating large-scale multimodal datasets
with long videos and aligned textual data is challenging and time
consuming, limiting the research conducted in this domain.

Previous work on video-to-video summarization identifies highlights
 from YouTube videos, TV shows, or
 movies~\cite{song2015tvsum,gygli2014creating,de2011vsumm,papalampidi2020movie}.
 However, in most cases, either the videos are short or the datasets
 are small with a few hundred examples. There is also limited work on
 video-to-text summarization. We are only aware of one large-scale
 multimodal dataset for this task, namely
 How2~\cite{sanabria2018how2}, which again contains short videos
 (i.e.,~2--3 minutes long) with simple semantics, and short,
 single-sentence summaries.

In this paper, we focus on video-to-text summarization and investigate
how to best utilize multimodal information for condensing long inputs
(e.g.,~an hour-long TV show) into long outputs (e.g.,~a multi-sentence
summary). We create a multimodal variant of
SummScreen~\cite{chen2021summscreen}, a recently released dataset
comprising of transcripts of TV episodes and their summaries. We
collect \mbox{full-length} videos for~4,575 episodes and multiple
reference summaries. We build our model on top of a pre-trained
sequence-to-sequence architecture
(i.e.,~BART;~\citealt{lewis2020bart}) fine-tuned on summarization and
capable of generating fluent long text. We convert its textual encoder
to a multimodal one by adding and tuning only adapter
layers~\cite{rebuffi2017learning,houlsby2019parameter}, which account
for 3.8\% of model parameters. We also explore strategies for
\emph{content selection}, since the input is too long to fit into
standard sequence-to-sequence models. Empirical results across 
evaluation metrics demonstrate that multimodal information yields
superior performance over just text, both in terms of content
selection and summarization; this is the case even when our
adapter model is compared to fully fine-tuned approaches and more
memory-heavy architectures (e.g.,~Longformer;
\citealt{Beltagy2020Longformer}) that can process the entire input. 

Our contributions can be summarized as follows: (1)~we augment
SummScreen \cite{chen2021summscreen} with multimodal information,
providing videos aligned with transcripts and summaries; to the best
of our knowledge, this constitutes the largest available resource for
long video multimodal summarization; (2)~we propose a \emph{parameter efficient}
approach to augment a pre-trained textual summarizer with
multimodal information; and (3)~explore different methods for
identifying salient moments in a long video and show that
multimodal information also improves content selection.


\footnotetext{TACoS contains only 127 cooking videos without corresponding transcripts and hence cannot be used for multimodal summarization.}

\section{Related Work}

\paragraph{Video Summarization} Much previous work has focused on
text-to-text or video-to-video summarization. We provide a
comprehensive categorization of existing datasets according to
input/output length and modality in
Table~\ref{tab:datasets_overview}. \textit{Multimodal abstractive
  summarization} (video-to-text) has attracted less attention, mainly
due to the difficulty of collecting large-scale datasets.
How2~\cite{sanabria2018how2} is the only publicly available benchmark
for this task, it includes short instructional videos with textual
transcripts and one-sentence summaries.  We generate
multiple-sentence summaries from long videos and their
transcripts. Previous approaches to multimodal summarization have
focused on various modality fusion methods with small RNN-based models
\cite{palaskar2019multimodal}. We take advantage of large pre-trained
LMs \cite{lewis2020bart,raffel2020exploring,radford2019language} for
generating fluent textual summaries.

Recent years have also witnessed increasing interest in multimodal video
captioning, a task related to multimodal summarization, which aims to
generate one-sentence descriptions for localized events in short
videos
\cite{xu2016msr,rohrbach2017movie,zhou2018towards,lei2020tvr}. Existing
methods  employ strong language-and-vision encoders with massive
pre-training
\cite{li2020hero,luo2020univl,xu2021vlm,lei2020mart,li2021value},
while the decoder is typically shallow and under-trained. Although
good at generating short descriptions, they cannot
maintain fluency in long outputs with rich
vocabularies. 

Realizing the importance of large LMs for generation, recent work has
focused on how to efficiently render pre-trained LMs
multimodal. Notably, \newcite{tsimpoukelli2021multimodal} convert a
pre-trained LM into an image captioning model, by giving images as
prompts and training only a vision encoder. \newcite{yu2021vision}
summarize How2 videos by augmenting BART-base with visual information
via a new cross-attention block added to every encoder layer. However,
their approach cannot easily scale to BART-large and beyond since they
add a large number of new parameters, while the dataset sizes are
relatively small, leading to over-fitting.

\paragraph{Dialogue Summarization} In the context of text-to-text
generation, dialogue summarization is challenging due to the
difficulty of fitting very long input into pre-trained
sequence-to-sequence models. Longformer \cite{Beltagy2020Longformer}
alleviates this by employing local self-attention in combination with
global tokens for reducing the computational overhead. Despite recent
attempts to make self-attention more efficient
\cite{kitaev2019reformer,tay2020sparse,zaheer2020big}, it is still
unclear whether it has an advantage over content selection with a
full-attention mechanism \cite{zhang2021exploratory,shaham2022scrolls}
for long dialogue summarization. \newcite{zhong2021dialoglm}
incorporate dialogue-specific objectives for pre-training
summarization models, while \newcite{zhang2021summ} follow a different
approach and hierarchically summarize the input chunk-by-chunk.

\paragraph{Parameter-efficient Tuning} Fine-tuning is a common
approach for transferring pre-trained models to different tasks or
domains \cite{howard2018universal}. It is customary to fine-tune all
the parameters of the pretrained model which, however, becomes
prohibitive as model size and number of tasks grow.  Recent work has
proposed parameter-efficient transfer learning methods which
fine-tune only a small number of \emph{additional} parameters. Two
popular approaches include \emph{adapter tuning}, where bottleneck layers
are added and tuned at every layer of the
model~\cite{rebuffi2017learning,houlsby2019parameter} and \emph{prompt
  tuning}, where (soft) prompts are prepended as part of the
input~\cite{brown2020language,li2021prefix}. In this work, following recent adapter-based approaches that efficiently convert LMs to vision-and-language models~\cite{sung2022vl}, we utilize
the former method for adapting a textual summarizer to our multimodal
setting and dialogue input format. 

\begin{table}[t]
\small
\centering
\begin{tabular}{@{}l@{~}r@{~}r@{~}r@{}}
\hline
Episodes & \multicolumn{3}{r@{}}{4,575} \\ \hline
\multicolumn{4}{c@{}}{Input: transcript + video + audio}  \\ \hline
Shots & \multicolumn{3}{r@{}}{1,048,024} \\
Shots/episode & \multicolumn{3}{r@{}}{193.64 (109.09)} \\
Utterances/episode & \multicolumn{3}{r@{}}{322.76 (116.52)} \\
Tokens/episode & \multicolumn{3}{r@{}}{5720.55 (2223.38)} \\
\hline
\multicolumn{4}{c}{Output: summaries}  \\ \hline
Summaries/episode &  & 1.53 &(0.79) \\
TVMegaSite/\#tokens & 4,280 &  395.69 &(275.84) \\ 
YouTube/\#tokens  & 334   &136.22 &(45.12) \\ 
IMDb/\#tokens & 946  & 111.21& (82.18) \\ 
tvdb/\#tokens & 1,454  &126.14 &(82.14) \\ \hline
Training (unique  input-output pairs)  & \multicolumn{3}{r@{}}{5,199} \\
Validation episodes & \multicolumn{3}{r@{}}{296} \\
Testing episodes & \multicolumn{3}{r@{}}{296} \\
\hline
\end{tabular}
\caption{SummScreen\textsuperscript{3D} statistics. For summaries, we
  show their provenance, number of summaries per site (second column),
  and  mean
  number of tokens per summary; standard deviations are shown in parentheses. }
\label{tab:new_dataset_statistics}
\end{table}

\section{The SummScreen\textsuperscript{3D} Dataset} \label{sec:dataset}

SummScreen~\cite{chen2021summscreen} is a long dialogue summarization
dataset containing transcripts from TV episodes and human-written
abstractive summaries\footnote{\url{https://github.com/mingdachen/SummScreen}}. We extend this dataset to a multimodal setting
by also considering the corresponding full-length videos. SummScreen
is divided into two subsets depending on the  series genre:
SummScreen-FD and SummScreen-TMS. We use the latter subset which
mostly covers soap operas from
TVMegaSite\footnote{\url{http://tvmegasite.net}}, as it is easier to
obtain full-length videos and each series has hundreds of episodes.

For each episode in SummScreen-TMS, we automatically search for the
title and release date in Youtube. If there is a match with large
duration (indicating that this is a full episode rather than a
segment), we download the video and closed captions (CC). Overall, we
collected videos for~4,575 episodes from five different shows in
SummScreen-TMS.\footnote{We will release scripts for data collection
  and processing.} In addition to TVMegaSite summaries (distributed
with SummScreen), we further retrieved summaries from YouTube
descriptions, IMDb, and tvdb, again using the episode title and
release date as search terms.  The statistics of our dataset which we
call SummScreen\textsuperscript{3D} (3D for language, video, and
audio) are in Table~\ref{tab:new_dataset_statistics} and we provide
further details in Appendix \ref{sec:appendix_dataset}. As can be
seen, each episode has (on average) multiple references which vary in
length (TVMegaSite summaries are  longest).

We split SummScreen\textsuperscript{3D} into training, validation, and
test sets with the same distribution over different shows per set. We
reserved 296 episodes for validation and the same number for testing,
and used the rest for training. Since we have multiple reference
summaries for some episodes, we increased the size of the training set
by adding~$m$ episode-summary pairs, matching the same episode with
each of its $m$~references. This resulted in 5,199 unique samples for
training.

\begin{figure*}[t]
    \tiny
    \centering
    \begin{subfigure}[b]{0.54\textwidth}   
        \tiny
        \centering 
        \hspace*{-.5cm}\includegraphics[width=1.05\textwidth,page=1]{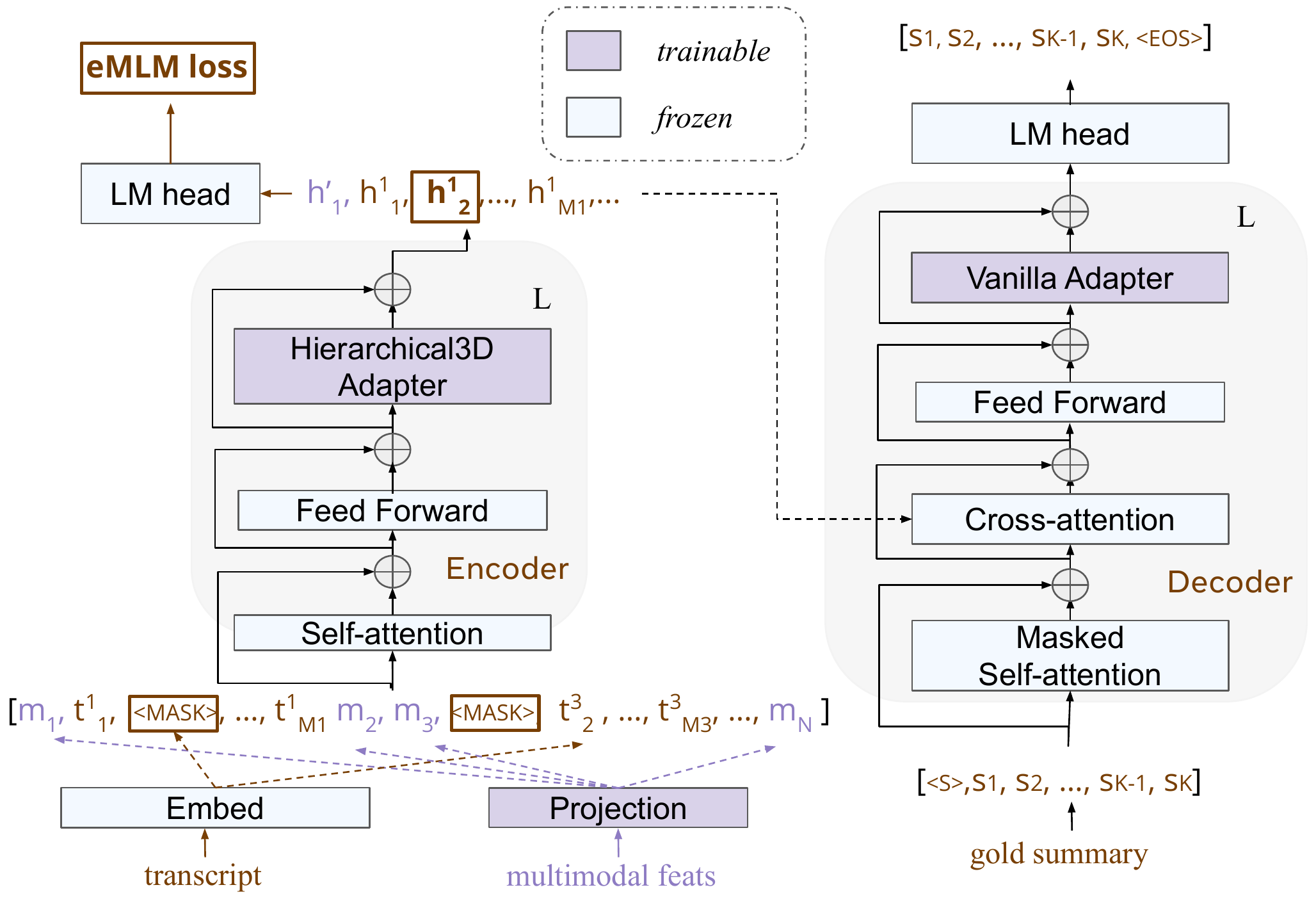}
        \caption[]%
        {{\small Multimodal augmentation of textual BART.}}    
        \label{fig:multimodal_augmentation}
    \end{subfigure}
    \begin{subfigure}[b]{0.44\textwidth}   
        \tiny
        \centering 
        \includegraphics[width=\textwidth,page=5]{model_figures_new.pdf}
        \caption[]%
        {{\small Hierarchical3D adapter for the encoder layers.}}    
        \label{fig:adapters}
    \end{subfigure}
    \vspace{1.5em}
    \caption[ ]
    {\small Multimodal augmentation of pre-trained BART. We augment
      the encoder and decoder layers with adapters which we fine-tune
      on the target dataset, while the remaining network is frozen. As
      input, we consider textual tokens and coarse-grained multimodal
      information which we prepend before each utterance. We also
      corrupt part of the textual input during training and add an
      auxiliary MLM loss to the encoder for predicting the corrupted
      tokens. On the right, we show the hierarchical adapter added to
      each encoder layer: after down-projecting all representations,
      we only consider the multimodal ones and further contextualize
      them via attention. Then, we combine the representations and
      up-project again to the original model dimension.}
    \label{fig:model}
\end{figure*}

\section{Video-to-Text Summarization}

Our approach leverages the generation capabilities of large
pre-trained sequence-to-sequence models
\cite{lewis2020bart,raffel2020exploring}. As our backbone model, we
employ BART-large~\cite{lewis2020bart} which has been fine-tuned on
\mbox{CNN-DailyMail}~\cite{nallapati2016abstractive,zhang2021exploratory} and
has thus acquired a summarization inductive bias.
As TV show transcripts are very long and cannot fit into BART, we
select a subset of utterances (i.e.,~speaker turns) as input via content selection (see details in Section~\ref{sec:preselection}). We transfer this model to our
task and domain (i.e.,~multimodal dialogue summarization), by adding
adapter layers \cite{rebuffi2017learning,houlsby2019parameter} in both
the encoder and decoder, and tuning them on
SummScreen\textsuperscript{3D} while keeping the rest of the network
frozen. We briefly discuss below our backbone text-based model and
then elaborate on how we incorporate multimodal information.

\subsection{Backbone Textual Model}

Our summarizer follows a standard sequence-to-sequence Transformer
architecture \cite{NIPS2017_7181}. The encoder maps tokens $[t_1, t_2,
\dots, t_N]$ to a sequence of contextualized representations $[h_1,
h_2, \dots, h_N]$ which are then fed to the decoder for generating the
summary. The encoder consists of~$L$ stacked layers, each of which has
a self-attention block for contextualizing the token representations,
followed by a feed-forward network. The decoder has a similar
architecture, it additionally contains a \emph{cross-attention} block
for identifying relations between the input and currently generated
text and makes use of \emph{masked} self-attention to control 
access to context for each token. The decoder is followed by a linear
layer (i.e.,~Language Model (LM) head) which projects the output
representations onto the vocabulary and a final softmax layer. The
model is optimized for predicting the next token~$s_{t+1}$ in the
summary given $[s_0, s_1, \dots, s_t]$, the context generated so far,
and the transcript $[t_1, t_2, \dots, t_N]$.

\subsection{Multimodal Augmentation} \label{sec:multimodal_augmentation}

Our hypothesis is that adding multimodal information to a textual
summarizer (i.e.,~converting the textual encoder to a multimodal one)
will increase the quality of its output summaries. We expect that the
video/audio will compensate for important non-verbal information
typically absent from the transcript (e.g.,~who is speaking to whom,
who is present in the same room, who is crying or yelling).  We
further expect multimodal information to make up for the loss of
context incurred by content selection.  We next describe how we
compute multimodal representations for an episode and how we augment
BART with these representations.

\paragraph{Multimodal Representations} We use \emph{utterances} as the
unit of representation for multimodal information. We segment episodes
into shots (using
PySceneDetect\footnote{\href{https://github.com/Breakthrough/PySceneDetect}{https://github.com/Breakthrough/PySceneDetect}})
and map these to utterances in the corresponding
transcript. Specifically, we align the closed captions in the video
which are time-stamped to the utterances in the transcript using
Dynamic Time Warping (DTW;
\citealt{myers1981comparative,papalampidi2020movie}). We thus create a
one-to-many alignment where an utterance corresponds to one or more
shots.  For each shot, we extract textual, visual, and audio features
(see Appendix~\ref{sec:appendix_preprocessing} for details), and
compute an utterance-level representation for each modality by average
pooling over all aligned shots.

Given textual~$x_i$, visual~$v_i$, and audio~$a_i$ representations for
utterance~$i$, we learn a multimodal representation as part of our
network:
\begin{equation}
\label{eq:proj_2}
  \begin{gathered}
\hspace*{-.1cm}    x'_i\hspace*{-.3ex}=\hspace*{-.3ex}f(W_x x_i)~~~ 
    v'_i \hspace*{-.3ex}=\hspace*{-.3ex} f(W_v v_i)~~~ 
    a'_i \hspace*{-.3ex}=\hspace*{-.3ex} f(W_a a_i) \\ 
    m_i = f(W_m [x'_i;v'_i;a'_i]) 
\end{gathered}
\end{equation}
where $f(\cdot)$ is the ReLU activation function,
$[\cdot;\cdot;\cdot]$ denotes concatenation, $W_x\in{\rm I\!R}^{d_x
  \mathrm{x} d_i}, W_v\in{\rm I\!R}^{d_v \mathrm{x} d_i}, W_a \in{\rm
  I\!R}^{d_a \mathrm{x} d_i}$, and $W_m \in{\rm I\!R}^{3d_i \mathrm{x}
  d_m}$ are learnable matrices; $d_i$ and $d_m$ are the input and
model dimensions with $d_i << d_m$, and $m_i$ is the final multimodal
representation corresponding to the $i^{th}$ utterance in the
transcript.

\paragraph{Multimodal Encoder} In order to integrate utterance-level
multimodal representations with BART, we consider a ``global utterance
token'' inspired by the Longformer architecture
\cite{Beltagy2020Longformer}. We preprocess the input into utterances
and prepend a global token $<$EOS$>$ per utterance as a placehoder for multimodal representations. The encoder thus receives as input sequence
$[{\bf{m_1}},t^{1}_1, t^{1}_2, \dots, t^{1}_{M_1}, \dots, {\bf{m_N}},
t^{N}_1, t^{N}_2, \dots, t^{N}_{M_N}]$ where, ``global''
representations $\bf{m}$ constitute a rich multimodal space
(i.e.,~they are not learned solely from text via local
self-attention). We illustrate this in
Figure~\ref{fig:multimodal_augmentation}.

\subsection{Self-supervised Auxiliary Guidance} \label{sec:MLM_objective}

Our primary loss for training the model described above is the
negative log likelihood of predicting the next token in the summary
given  episode~$\mathcal{E}$:
\vspace{-0.6em}
\begin{gather}
  L_{LM} = \frac{1}{K}\sum_{t \in  [1,K]} - \log p(s_t|s<t;\mathcal{E})
  \end{gather}
 \vspace{-0.1em}
 We further wish to encourage the model to attend to multimodal
  information and learn a meaningful projection
  (Equation~\eqref{eq:proj_2}). To do this, we corrupt part of the
  textual input by masking tokens (see bottom left part of
  Figure~\ref{fig:multimodal_augmentation}) and adding an auxiliary
  masked language modeling (MLM) loss for the initial training steps
  only. So as not to disrupt the bias of the decoder, which is already
  trained on textual summarization, we apply the MLM loss in the
  outputs of the encoder while the model is trained on the downstream
  task. Given token-level encoder outputs~$[h_1, h_2, \dots, h_N]$, we
  copy and re-use the LM head of the decoder in order to project them
  into the vocabulary (see top left part of
  Figure~\ref{fig:multimodal_augmentation}). And compute the negative
  log likelihood only for the set of masked tokens $\mathcal{M}$:
  \vspace{-0.6em}
\begin{gather}
\mathcal{L}_{eMLM} = \frac{1}{|\mathcal{M}|}\sum_{t
  \in \mathcal{M}} - \log p(t|h_{t_i\notin \mathcal{M}})
\end{gather}
\vspace{-0.1em} We refer to this loss as encoder-based MLM loss (eMLM;
\citealt{baziotis2021exploring}). It trains the encoder to reconstruct
input text representations while attending to multimodal
information. After $X$~initial training steps, we drop the auxiliary
loss and stop corrupting the textual input in order for the model to
be optimized on summarization.  We use a mixture of content word
corruption (i.e.,~masking out named entities, nouns, and verbs
excluding auxiliaries) and whole utterance corruption
\cite{zhang2020pegasus,zhong2021dialoglm}. We provide more details in
Section~\ref{sec:training_details}.

\subsection{Hierarchical3D Adapters} \label{sec:hierarchical_adapters}
We specialize BART for our multimodal summarization task by inserting
adapter modules \cite{rebuffi2017learning,houlsby2019parameter} into
each encoder and decoder layer (after the feed-forward block). Each
adapter adds only a small number of new parameters, which are randomly
initialized and tuned on our end task, while the rest of the network
is frozen. A vanilla adapter takes as input hidden representations
${\bf{[u_1}}, h^{1}_1, h^{1}_2, \dots, {\bf{u_N}},\dots, h^{N}_{M_N}]$, where $h^{1}_1, h^{1}_2, \dots, h^{N}_{M_N}$ are textual token-level hidden representations and $\bf{u_1}, \dots, \bf{u_N}$ are multimodal utterance-level hidden representations (in accordance to the input format presented in Figure~\ref{fig:multimodal_augmentation}),
and performs the following transformations:
\vspace{-0.6em}
\begin{gather}
    h_{down,i} = f(\operatorname{LN}(W_dh_i+b_d)) \\
    h_{up,i} = W_uh_{down,i} + b_u \quad
    h_i = h_i + h_{up,i}\label{eq:up_project}
\end{gather}
where $W_d\in{\rm I\!R}^{d_m \mathrm{x} d_{\mathcal{B}}}$, $d_m$ is
the model dimension, $d_{\mathcal{B}}$ is the bottleneck dimension of
the adapter, $f(\cdot)$ is a non-linearity, $\operatorname{LN}$ a
trainable layer normalization, $W_u \in {\rm I\!R}^{d_{\mathcal{B}}
  \mathrm{x} d_m}$, $b_d$, and $b_u$ are the corresponding bias
vectors, and $h_{down,i}$ and $h_{up,i}$ are down and up projections of~$h_i$.

In this work, we augment the vanilla adapters of the \textit{encoder}
with a hierarchical structure (illustrated in
Figure~\ref{fig:adapters}). After computing (low level) self-attention
between all input \textit{textual tokens} in an encoder layer, we add
a hierarchical adapter to compute \textit{higher-level interactions}
between \textit{utterance-level multimodal} representations. By
including this interaction block in the adapter, we can better
propagate long-range dependencies between utterances and enforce a a
more global view of the events in an episode and their associations,
while keeping the number of trainable parameters low.

We first compute interaction (aka similarity) matrix~$H$ between
utterances (see Figure~\ref{fig:adapters}) based on their
\textit{multimodal representations} $[m_1, m_2, \dots, m_N]$ using the
scaled dot product: \vspace{-0.4em}
\begin{gather}
e_{ij} = (W_im_i+b_i)(W_jm_j+b_j)/\sqrt{d_m}
\end{gather}
where $W_i, W_j$ are learnable projection matrices, $d_m$ is the model
dimension, and $e_{ij}$ is the degree of similarity between $m_i$ and
$m_j$. 

At each adapter layer of the encoder, after down-projecting all
vectors to the bottleneck dimension, we further contextualize
utterance-level multimodal representations~$u_{down,i}$ with respect
to each other given the degree of similarity provided by~$H$
(''Contextualize'' block in Figure~\ref{fig:adapters}):
\vspace{-0.6em}
\begin{gather}
    u'_{down,i} = \sum_{k=1}^N \mathrm{r}(H_{ik}/\tau)u_{down,k} + u_{down,i}
\end{gather}
where $N$ is the number of utterances, $\mathrm{r}(\cdot)$ is the softmax function, and $\tau$ is
a low temperature parameter ($<1$) for increasing sparsity.  After
contextualization, we up-project all vectors to the original dimension~$d_m$, as in vanilla adapters
(Equation~\eqref{eq:up_project}).

\section{Content Selection} \label{sec:preselection}

As explained earlier, episodes in SummScreen\textsuperscript{3D} are
very long ($\sim$5,720 tokens). BART, which has a maximum token length
of~1,024, can approximately encode one fifth of the
transcript.\footnote{We can extend positional embeddings to~1,536 by
  applying bilinear interpolation, however, the memory requirements
  would be prohibitive for longer sequences.} We therefore perform
content selection, i.e.,~identify salient utterances and give these as
input to BART. We describe below three approaches inspired by
information retrieval, summarization
\cite{gehrmann-etal-2018-bottom,liu-lapata-2019-hierarchical}, and
computational narrative analysis
\cite{papalampidi2020movie,papalampidi2021film}.

\vspace{-0.5em}

\paragraph{Retrieval-based Selection} We follow previous approaches~\cite{zhang2021exploratory} in determining salient content with
BM25~\cite{robertson2009probabilistic}. BM25 is a widely known
retrieval model similar to tf*idf. It assigns each utterance a
``relevance'' score (by comparing it against the entire
transcript). Utterances with high scores are deemed salient and the $K$ best ones are selected.

\vspace{-0.5em}

\paragraph{Learning-based Selection} Alternatively, we may also model
content selection as a binary classification problem.  Given a
transcript containing $N$~utterances we predict whether each should be
selected as input for the downstream summarization task (label~1) or
not (label~0). We create noisy labels by matching transcript
utterances to (reference) summary sentences. Specifically, we encode
sentences and utterances via Sentence-BERT \cite{reimers2019sentence},
and assign a positive label to the utterances most similar to the
reference sentences.  A content selector is then trained on these
pseudo-labels to identify salient utterances.  We can also incorporate
multimodal information in this content selection setting, using the
same utterance-level representations fed into BART. We first
contextualize them via a shallow transformer encoder, and add a
classification head for predicting important utterances. The model is
optimized with binary cross-entropy loss. During inference we select
the top $K$~predicted utterances. 

\vspace{-0.5em}

\paragraph{Turning Point (TP) identification} We also perform content
selection based on a Turning Point (TP) identification
model~\cite{papalampidi2020movie,papalampidi2021film} pre-trained on
the TRIPOD movie dataset~\cite{papalampidi2019movie}. TPs are key
events in narratives; they are distinguished into five different types
depending on their functionality (e.g.,~Opportunity, Change of Plans,
Point of No Return, Major Setback, Climax).  The TP identification
model considers the same multimodal information as the content
selector above and identifies utterances that represent each TP. We
consider the top~$K/5$ predicted utterances per turning point.

\section{Experimental Setup}

\paragraph{Implementation Details} \label{sec:training_details}

We provide details of the multimodal feature
extraction (i.e.,~utterance-level visual, audio, and textual features) in Appendix~\ref{sec:appendix_preprocessing}.  We corrupt
the textual input and use the auxiliary eMLM loss
(Section~\ref{sec:MLM_objective}) only for the first $X=$1,500
training steps; we train our model for a total of 12,000
steps. During corruption, we mask out all content words (i.e.,~named
entities, verbs, and nouns) and a random 10\% of the input
utterances. For generating summaries during inference, we use beam
search with $\mathrm{beam}=5$ and 3-gram
blocking~\cite{paulus2018deep}. We provide further implementation details in Appendix~\ref{sec:appendix_training_details}.

\vspace{-.2cm}
\paragraph{Training vs. Inference} Although we experiment with
different content selection methods during inference, we randomly
sample input utterances during training. Random sampling acts as data
augmentation, since the model sees slightly different input-output
pairs during training at different iterations. We experimentally
verify in Section~\ref{sec:results} this is preferable to a fixed
selection of utterances, especially considering the small dataset
size. We select $K=60$ utterances to feed into BART models given the
input length limit, and order them according to their original
position in the transcript.

\begin{table}[t]
\small
\centering
\begin{tabular}{l|cccc} \hline
\multicolumn{1}{l|}{Selection} & {R-1} & {R-2} & {R-L}  \\ \hline 
Lead & 32.91 & 6.51 & 30.72 \\
Last & 32.65 & 6.41 & 30.59 \\
Middle & 33.02 & 6.70 & 31.03 \\
Random selection & 33.07 & 6.54 & 30.91 \\
\hspace{1em} $+$ Hierarchical3D & \underline{34.33} & 7.24 & 32.15 \\
Retrieval & 32.36 & 6.30 & 30.20 \\
\hspace{1em} $+$ Hierarchical3D & 33.83 & 6.89 & 31.42 \\
TP identification & 33.31 & 6.78 & 31.24 \\
\hspace{1em} $+$ Hierarchical3D & \underline{34.49} & 7.36 & 32.01 \\
Content Selection & 33.27 & 6.74 & 31.22 \\
\hspace{1em} $+$ Hierarchical3D & \textbf{34.51} & \textbf{7.62} & \textbf{32.64} \\
\hdashline
Pseudo-oracle & \textit{35.09} & \textit{7.96} & \textit{32.85} \\
\hspace{1em} $+$ Hierarchical3D & \textit{35.69} & \textit{8.42} & \textit{33.40} \\
\hline
\end{tabular}
\caption{Content selection methods for textual and multimodal BART
  ($+$Hierarchical3D).}
\label{tab:generation_results_summscreen_test_content_selection}
\end{table}

\vspace{-0.3em}

\paragraph{Evaluation Metrics} \label{sec:evaluation_metrics}

We evaluate the generated summaries using ROUGE F1~\cite{lin2004rouge}
against reference summaries\footnote{\url{https://pypi.org/project/py-rouge/}}. However, there is mounting evidence that
ROUGE is not always a good indicator of summary quality and does not
discriminate between different error types, especially factuality-related ones\footnote{We also experimented with using BERTScore~\cite{zhang2019bertscore} as an extra evaluation metric, but we observe no significant difference in performance between any pair of models, so we excluded the metric from our results.}. We therefore consider additional metrics based
on Question-Answering (QA). We obtain questions based on the gold
summaries and evaluate whether the correct answers exist in the
generated summaries. We expect factual summaries to
correctly answer a higher percentage of questions.

As in prior
work~\cite{maynez2020faithfulness,kryscinski2020evaluating,honovich2021q2},
we automatically generate QA pairs against reference summaries. We
identify named entities and nouns using
spaCy~\cite{honnibal2017spacy}, and feed them as gold answers
alongside the summaries to a question generator. We discriminate
between named entities and nouns as answer types for measuring
factuality in event-entity associations and other attributes
pertaining to nouns. We used T5-base~\cite{raffel2020exploring} as our
question generator and
RoBERTa-base~\cite{DBLP:journals/corr/abs-1907-11692} as the QA system
for answering questions given system generated summaries as input
passages. Both were fine-tuned on SQuAD2.0~\cite{2016arXiv160605250R}.

We measure accuracy as the partial overlap between gold and predicted
answers for named entities. For nouns, we resort to textual entailment
in order to account for synonyms and paraphrases in the generated
summaries. We concatenate the question with gold or generated answer
and predict a score for the directional relation between them. If the score is above $0.5$, we consider the generated answer correct. We used 
BART-large~\cite{lewis2020bart} fine-tuned on the MultiNLI
corpus~\cite{N18-1101} as our entailment model. 

We created a test suite of gold QA pairs, by retaining only those that
can be answered correctly by the QA model given the reference
summaries~\cite{honovich2021q2}. We overall generated 2,513 questions
for named entities and 381 questions for nouns for the 296 episodes in
our test set. On average, we have 8.5 questions per episode for named
entities and 2.3~questions for nouns.

\section{Results} \label{sec:results}

\paragraph{Content Selection} 
Table~\ref{tab:generation_results_summscreen_test_content_selection}
compares how different approaches to content selection influence
summarization performance according to ROUGE~F1.  We compare some
simple baselines like selecting the Lead, Middle, and Last 60
utterances from the transcript as well as at Random. In addition, we
compare a text only summarizer against our Hierarchical3D model. As
can be seen, differences amongst content selection methods are
generally small.  BM25 performs worse than random whilst a multimodal
content selector trained on pseudo-labels performs overall best. As an
upper bound, we also report results with oracle labels as input
demonstrating that there is still room for improvement. Regardless of
how content is selected, we observe that our Hierarchical3D variant
significantly improves performance, and interestingly, the performance
gap is larger when the selection method is weaker (e.g.,~random
vs. pseudo-oracle). This indicates that to a certain extent multimodal
information makes up for suboptimal content selection.

\begin{table}[t]
\small
\centering
\begin{tabular}{lcccc}
\hline
Models & {R-1} & {R-2} & {R-L}  \\ \hline
LED FT & 33.53 & \textbf{7.60} & 31.77 \\ 
DialogLED FT & 32.66 & 7.38 & 31.12 \\
$\mathrm{Summ}^N$ FT & 24.71 & 4.42 & 22.61 \\ \hline\hline
BART FT & 32.61 & 6.94 & 30.83 \\ 
BART AT & 33.27 & 6.74 & 31.22 \\
BART AT $+$ Hierarchical3D & \textbf{34.51} & \textbf{7.62} & \textbf{32.64} \\
\hline
\end{tabular}
\caption{Comparison of our model (BART AT + H-3D) with
  text-only summarizers for long dialogue summarization. For all BART
  variants we perform content selection (FT: full fine-tuning, AT:
  adapter-tuning).} 
\label{tab:generation_results_summscreen_test}
\end{table}

\vspace{-0.3em}

\paragraph{Text vs. Multiple Modalities}
In Table~\ref{tab:generation_results_summscreen_test} we compare our multimodal model (with the best performing content selector)
against textual summarizers developed for processing long input or specifically for dialogue
summarization. These include Longformer (LED;
\citealt{Beltagy2020Longformer}) with full fine-tuning\footnote{Adding
  and tuning only adapter layers in LED gave us inferior performance
  by a large margin, indicating that adapting such a network is not
  straightforward. Hence, converting it into a multimodal version is also more challenging.}, a variant of LED pre-trained on dialogues
(DialogLED; \citealt{zhong2021dialoglm}), and
$\mathrm{Summ}^N$~\cite{zhang2021summ}, a two-stage hierarchical
approach for long dialogue summarization. We also present
text-only BART variants, with full fine-tuning (FT) and adapter-tuning
(AT)\footnote{We also considered video-to-text models trained on video captioning (e.g.,~HERO;~\citealt{li2020hero}) during initial experimentation, but we find that these models fail to produce fluent multiple-sentence textual outputs due to the shallow and under-trained decoder and cannot process long video sequences alongside their transcripts due to computational constraints.}.

As can be seen in the second block of
Table~\ref{tab:generation_results_summscreen_test}, tuning only the
adapter layers (BART AT) does not hurt performance compared to full
fine-tuning (BART FT), presumably due to the small dataset size. Addition of multimodal information with hierarchical adapters
(BART AT + Hierarchical3D) yields substantial ROUGE improvements.
Interestingly, our performance is superior to fully fine-tuned,
memory-heavy models like LED or DialogLED that process the entire
transcript as input. This suggests that representations from multiple
modalities are more informative and lead to higher performance
compared to efficient self-attention mechanisms. $\mathrm{Summ}^N$
performs demonstrably worse than all one-stage methods. 

\begin{table}[t]
\small
\centering
\begin{tabular}{@{}lcccc@{}}
\hline
 Models & \multicolumn{2}{c}{Acc (NEs)} & \multicolumn{2}{c}{Acc (NNs)} \\
 & {text} & {$+$H-3D}  & {text} & {$+$H-3D} \\\hline
Random selection & 20.25 & 23.64 & 33.86 & 38.06 \\
TP identification & 21.65 & 24.07 & 40.42 & \textbf{40.68} \\
Content Selection & 20.65 & \textbf{24.71} & 38.58 & 39.37 \\
\hdashline
Pseudo-oracle & \textit{28.53} & \textit{29.64} & \textit{41.73} & \textit{42.00} \\
\hline \hline
LED FT & 20.89 & --- & 37.95 & --- \\
DialogLED FT & 21.09 & ---  & 36.22 & ---  \\
$\mathrm{Summ}^N$ FT & 18.03 & ---  & 34.91 & ---  \\
\hline
\end{tabular}
\caption{QA evaluation on named entities and nouns. We denote our Hierarchical3D model with H-3D.}
\label{tab:automatic_qa_test_set}
\end{table}

\vspace{-0.3em}

\paragraph{QA Evaluation} The results of our automatic QA evaluation
are summarized in Table~\ref{tab:automatic_qa_test_set}. The first
block focuses on model performance with different content selection
variants. We only compare text-only and multimodal ($+$H-3D)
BART. Again, we find that augmenting BART with multimodal information
regardless of the selection method improves accuracy, especially for
named entities (Columns 2 and 4 in
Table~\ref{tab:automatic_qa_test_set} vs 3 and 5). This is true even
when content is selected by a pseudo-oracle suggesting that multimodal
information provides better associations between events and entities,
even when the input contains all salient information.  We further
observe that supervised content selection and TP identification offer
the best performance. The second block compares our approach with
state-of-the-art models on dialogue summarization; we find these
models perform on par or slightly worse than textual BART (depending
on the content selection method) which casts doubts on their ability
to efficiently consume longer inputs. We report more entity-specific
metrics in Appendix~\ref{sec:appendix_entity_specific_eval} and
present examples of generated summaries and QA pairs in
Appendix~\ref{sec:appendix_examples}.

\begin{table}[t]
\small
\centering
\begin{tabular}{lrrr}
\hline
Modality & {R-1} & {R-2} & {R-L} \\ \hline
Text & 34.74 & 7.11 & 32.46 \\
Audio & 33.95 & 6.92 & 31.90 \\
Video & 34.86 & 7.24 & 32.73 \\
Multimodal & \textbf{34.95} & \textbf{7.51} & \textbf{33.01} \\
\hspace{1em} w/ vanilla adapters & 34.25 & 7.45 & 32.41 \\
\hspace{1em} w/o eMLM loss &  33.80 & 6.84 & 31.88 \\
\hspace{1em} w/o random augmentation & 33.45 & 6.48 & 31.81 \\
\hline
\end{tabular}
\caption{Role of multimodal information and hierarchical adapters
  (validation set).}
\label{tab:multimodal_ablation}
\end{table}

\vspace{-0.3em}

\paragraph{Ablation Studies}
In Table~\ref{tab:multimodal_ablation} we summarize our findings from
ablation studies which aim to isolate which modeling components
contribute to better performance. We observe that individual
modalities (Text, Audio, Video) perform worse on their own than in
combination (Multimodal). The least informative modality is audio,
while the most informative one is video. While considering multimodal
information, we substitute the hierarchical adapters in the encoder
with vanilla adapters and observe a small drop in performance. Removal
of the auxiliary eMLM loss during training leads to a further
performance drop. The auxiliary loss is crucial rendering the
textual encoder  multimodal and forcing an already tuned summarizer
to consider a different type of input. Finally, data augmentation (via
random content selection) during training is also important given the
small size of the dataset and BART encoder length restrictions.  We
also report additional ablation experiments in
Appendix~\ref{sec:appendix_ablation} which show similar trends for the
contribution of different modalities to content selection.

\section{Conclusions}

In this work, we addressed the task of multimodal abstractive
summarization and created SummScreen\textsuperscript{3D}, a new
dataset which we hope will facilitate future research in this
direction. We incorporated multimodal information into a pre-trained
textual summarizer in a parameter-efficient manner and have
experimentally shown performance gains over text-only models.  Our
experimental results further underscore the importance of (multimodal)
content selection compared to approaches focusing on self-attention
variants for long dialogue summarization. In the future, we plan to
explore more \emph{structure-aware} representations for \emph{all} input
modalities in order to improve factuality (e.g.,~entity-event
associations) in the generated summaries.

\section{Limitations}

Our approach considers only coarse-grained (i.e.,~utterance-level)
multimodal information which we demonstrate is beneficial for
summarization.  More detailed frame-level visual information
e.g.,~person identification and object recognition in frames, would be
useful.  However, considering frame-level representations for
hour-long videos would bring a considerable increase in memory
requirements and additional difficulties in aligning different
modalities (e.g.,~frames vs. tokens vs. audio segments). We leave
these challenges to future work and believe that structure-aware
methods are necessary for addressing the current limitations.

Following previous
work~\cite{maynez2020faithfulness,kryscinski2020evaluating,honovich2021q2},
we advocate the use of automatic QA-based methods for evaluating the
generated summaries. However, more analysis and experimentation is
necessary for understanding whether these methods are trustworthy
(e.g.,~\citealt{tang2022understanding}), especially considering that
human evaluation is difficult if not infeasible for long dialogue
summarization.

\bibliography{anthology,custom}
\bibliographystyle{acl_natbib}

\newpage
\appendix

\begin{table}[t]
\small
\centering
\begin{tabular}{lr}
\hline
As The World Turns (atwt) & 1356 \\
Bold and the Beautiful (bb) & 1113 \\
Guiding Light (gl) & 836 \\
One Life to Live (oltl) & 1118 \\
Port Charles (pc) & 501 \\
\hline
\end{tabular}
\caption{Distribution of different TV shows in SummScreen\textsuperscript{3D}.}
\label{tab:new_dataset_statistics_2}
\end{table}

\section{Dataset Details}
\label{sec:appendix_dataset}

As mentioned in Section~\ref{sec:dataset}, we create a multimodal version of the SummScreen dataset~\cite{chen2021summscreen} by collecting the full-length videos of the episodes contained in the original dataset. Overall, we retrieved videos from YouTube for five different TV shows (i.e.,~soap operas). We present in Table~\ref{tab:new_dataset_statistics_2} the names of the TV shows and the number of episodes per show. We made sure to have enough episodes from each TV show and maintain the same distribution when splitting the dataset into train, validation, and test.  

\section{Implementation Details}

\subsection{Dataset Pre-processing} \label{sec:appendix_preprocessing}

Given full-length video, we extract features for all modalities at the
utterance-level as mentioned in
Section~\ref{sec:multimodal_augmentation}.  For text, we extract
sentence-level features using Sentence-BERT
\cite{reimers2019sentence}. Each utterance in the transcript is thus
represented by a fixed-size vector. For the frames, we extract two
types of features: frame-level features using the CLIP visual
encoder~\cite{radford2021learning} and motion-level features from
video clips using Slowfast~\cite{feichtenhofer2019slowfast}. We then
aggregate frame- and motion-level features to utterance-level given
the automatic alignment by mean pooling. Finally, for audio, we use
YAMNet pre-trained on the AudioSet-YouTube corpus
\cite{gemmeke2017audio} for classifying audio segments into 521 audio
classes (e.g.,~tools, music, explosion); for each audio segment
contained in a shot, we extract features from the penultimate layer,
and then aggregate representations again to utterance-level via mean
pooling.

\subsection{Training Details} \label{sec:appendix_training_details}

We used the Adam algorithm~\cite{kingma2015adam} for optimizing our
networks.  We trained all models with a learning rate of $3e{-5}$ for
12k steps using a linear warm-up of 500 steps, followed by inverted
squared decay. All BART-based models were trained with batch size of 1
episode on 4 P100 GPUs with 16GB memory and label smoothing
~\cite{szegedy2016rethinking} of 0.1. To fine-tune the LED-based
models, we used 4 A100 GPUs with 80GB memory. It took approximately 12 hours to fully train each of these models. Fully fine-tuned models have 406M parameters, which are all fine-tuned on the target dataset, whereas our multimodal adapter-augmented model has 421.6M parameters, from which we only train 15.6M parameters (i.e.,~multimodal projection layer and adapter layers) on the target dataset.  This means that we only tune $\sim$3.8\% of the model parameters of the fully fine-tuned models.

\begin{table}[t]
\small
\centering
\begin{tabular}{lrrr}
\hline
 & \textbf{Pre (\%)} & \textbf{Re (\%)} & \textbf{F1 (\%)} \\ 
\hline
\multicolumn{4}{c}{Unsupervised}  \\ \hline
Random & 19.55 & 20.90 & 20.06 \\
Retrieval & 24.63 & 26.62 & 25.40 \\
TP identification & 20.35 & 22.10 & 21.04 \\
\hline
\multicolumn{4}{c}{Supervised}  \\ \hline
Multimodal & \textbf{47.57} & \textbf{50.68} & \textbf{48.57} \\
Text & 45.26 & 48.54 & 46.52 \\
Vision & 22.97 & 24.91 & 23.73 \\
Audio & 21.54 & 23.29 & 22.23 \\
\hline
\end{tabular}
\caption{Role of multimodal information in content selection. Precision (Pre), Recall (Re) and F1 for selecting important utterances from the episode. The supervised models are trained given the pseudo-oracle labels.}
\label{tab:content_selection_ablation}
\end{table}

\section{Additional Experimental Results}
\label{sec:appendix_results}

\subsection{Ablation Study on Content Selection} \label{sec:appendix_ablation}

In Table~\ref{tab:content_selection_ablation}, we directly test the
performance of different content selectors. We report precision (Pre),
recall (Re), and F1 score of model variants based on pseudo-oracle
labels. We first consider selectors which have not been trained with
pseudo-oracle labels, such as Random, Retrieval (i.e.,~BM25) and TP
identification (we refer to these approaches as unsupervised). We
observe that unsupervised baselines have significantly lower F1 score
in comparison with a supervised approach. Interestingly, although TP
identification ``agrees less'' with the pseudo-oracle labels in
comparison with BM25, TPs still present competitive performance
against the supervised content selector on abstractive textual
summarization (e.g.,~Table~\ref{tab:automatic_qa_test_set}). Finally,
comparing the multimodal supervised content selector with equivalent
unimodal models, we observe that the highest performance is achieved
by combining all modalities. Investigating the unimodal variants, we
also conclude that the most informative one is the textual modality,
while using visual or audio cues alone is not enough to predict
salient content.

\begin{table*}[t]
\small
\centering
\begin{tabular}{lrrrrrr}
\hline
 & BoC-p & BoC-r & BoC-f1 & BoR-p & BoR-r  & BoR-f1 \\ \hline
Random selection & 82.55 & 38.71 & 52.71 & 29.82 & 9.39 & 14.28 \\
\hspace{1em} + Hierarchical3D & 81.80 & 47.37 & 60.00 & 31.75 & 13.77 & 19.21 \\
TP identification & \textbf{84.31} & 38.93 & 53.26 & \textbf{36.79} & 10.33 & 16.13 \\
\hspace{1em} + Hierarchical3D & 82.20 & 47.10 & 59.89 & 34.82 & \textbf{14.10} & \textbf{20.07} \\
Content Selection & 81.60 & 36.59 & 50.52 & 30.54 & 8.58 & 13.40 \\
\hspace{1em} + Hierarchical3D & 81.90 & \textbf{48.48} & \textbf{60.91} & 33.04 & \textbf{14.37} &\textbf{ 20.03} \\
\hdashline
Pseudo-oracle & \textit{87.42} & \textit{46.95} & \textit{61.09} & \textit{37.92} & \textit{14.40} & \textit{20.87}  \\
\hspace{1em} + Hierarchical3D & \textit{85.53} & \textit{52.37} & \textit{64.96} & \textit{36.67} & \textit{17.51} & \textit{23.70} \\
\hline
LED FT & 82.28 & 33.54 & 47.65 & 34.35 & 10.64 & 16.25 \\
DialogLED FT & 82.93 & 38.19 & 52.27 & 31.71 & 10.32 & 15.57 \\
$\mathrm{Summ}^N$ FT & 82.74 & 29.14 & 43.10 & 34.73 & 9.39 & 14.78 \\
\hline
\end{tabular}
\caption{Entity-specific metrics (test set).}
\label{tab:entities}
\end{table*}

\subsection{Entity-specific Evaluation} \label{sec:appendix_entity_specific_eval}

\newcite{chen2021summscreen} propose a set of entity-specific metrics
in order to investigate the role of characters, which are fundamental
in TV shows, in the generated summaries. Specifically, they measure
several bag of character (BoC) metrics based on character overalp
between generated and gold standard summaries. They define precision
as the fraction of the correctly mentioned characters with respect to
all characters that appear in the generated summary (BoC-p) and recall
as the fraction of the correctly mentioned characters with respect to
all characters that appear in the gold summary (BoC-r). Given
precision and recall, we also measure F1-score (BoC-f1).

Apart from correctly mentioned characters,
\newcite{chen2021summscreen} also compute similar bag of words metrics
for the relations between characters in the summaries. Specifically,
they consider a pair of characters related if they appear in the same
sentence in the summary. They do not account for the direction of the
relations and focus only on co-occurrence.  For measuring these
relations, they again consider precision (BoR-p) and recall (BoR-r) of
the intersection of pairs of characters similarly to computing the BoC
metrics. We also report F1-score (BoR-f1), given the precision and
recall for character relations.

We summarize the entity-specific results in
Table~\ref{tab:entities}. Overall, especially when considerring the F1
scores for characters and relations, we arrive to similar conclusions
as with our automatic QA evaluation
(Table~\ref{tab:automatic_qa_test_set}). First, the multimodal
information that is incorporated in our Hierarchical3D approach
increases most entity-specific metrics in comparison with the
text-only variants. Regarding different content selection methods, TP
identification and supervised content selection again perform best in
comparison with random selection, although differences are not
large. Finally, we achieve the best F1 scores in both entity- and
relation-specific metrics by using oracle selection, indicating that
there is still room for improvement. Interestingly, we again observe a
further increase in performance by adding multimodal information in
the pseudo-oracle variant, suggesting that video-based information is
important even when we consider the most salient parts of an episode.

Finally, we also compare our approach with state-of-the-art, fully
finetuned textual summarizers for long dialogues. We again notice that
$\mathrm{Summ}^N$ is the weakest option regarding the entity-specific
metrics. Next, considering efficient architectures for modeling the
entire input (i.e.,~LED, DialogLED) has competitive performance with
our text-only variants with content selection. However, Hierarchical3D
that considers multimodal information outperforms these memory-heavy
models while training only a small fraction of model parameters. This
further validates our hypothesis that the video can provide further
information that is more important for high-quality summaries than
exploring efficient methods to process the entire textual input.

\begin{table*}[t]
\small
\centering
\begin{tabular}{L{24em}L{15em}l}\hline
\multicolumn{1}{c}{\textbf{Summary}} & \multicolumn{1}{c}{\textbf{Question}} & \multicolumn{1}{c}{\textbf{Answer}} \\ 
\hline \hline
\multirow{2}{24em}{Sage goes to live with Jack after she learns Carly is planning to marry Craig. Meg agrees to marry Dusty.} & Who does Meg agree to marry? & Dusty \\
& Who does Sage go to live with? & Jack \\ \hline\hline
\multirow{2}{24em}{Joshua is busy preparing for Allison's arrival, as he unveils Kevin's latest creation; a portrait of Allison and Joshua in their wedding attire. Lucy goes to church to plead for answers. Ian overhears her plea and swears that he will not let her die. Livvie shows Joshua a picture of Allison appearing to be dead and tells him that he was right her fangs are poisoned.} & Who goes to church to plead for answers? & Lucy \\
 & Who swears he will not let Lucy die? & Ian \\
 & What does Lucy do at church? & plea \\
 & What part of Allison's body is poisoned? & flangs \\ \\
\multicolumn{3}{c}{}\\\hline \hline
 \multirow{2}{24em}{Lizzie and Jonathan spend some time with their baby. Jonathan gives in to one of Alan s demands. Gus and Harley find a disk with some interesting information on it. Gus still can t figure out what it is that Blake has on him. Dinah and Mallet argue over who will be the next WSPR star. Tammy is heartbroken after a visit to the hospital. Jonathan and Lizzie find out their baby has a medical condition, and make a run for it. Alan realizes that he may have been outwitted by Jonathan. Gus vows to get to the bottom of his supposed secret.} & What does Gus vow to find out about Blake? & secret \\
 &  What is Lizzie and Jonathan spending time with? & baby \\
 & What do Gus and Harley find? & disk \\
 & What do Lizzie and Jonathan do when they learn their baby has a medical condition? & run \\\\\\\\
\hline
\end{tabular}
\caption{Examples of summaries and automatically generated QA pairs.}
\label{tab:examples_QA}
\end{table*}

\begin{table*}[t]
\small
\centering
\begin{tabular}{L{5em}L{40em}}
\hline
\multicolumn{1}{l}{\textbf{Model}} & \multicolumn{1}{c}{\textbf{Summary}} \\ 
\hline
\textbf{Gold} & Caleb is upset when Livvie tells him that Joshua has the ring.  Joshua attempts to sway Ian to the dark side, but Ian vows he will
 continue to fight Joshua and the other vampires.  Rafe tells Caleb the only way he can defeat Joshua now is to remain human and Livvie reluctantly agrees.  Lucy pleads with Victor to fight Joshua, however, it s too late, as Victor tells her he enjoys the power Joshua has given him. Karen realizes Frank is a vampire. \\ 
\multirow{7}{5em}{\textbf{QA pairs}} & \tabitem Who tells Lucy that he enjoys the power Joshua has given him? -\color{tp2}{\textbf{Victor}} \\
& \tabitem Who does Karen realize is a vampire? -\color{tp2}{\textbf{Frank}} \\
& \tabitem Who pleads with Victor to fight Joshua? -\color{tp2}{\textbf{Lucy}} \\
& \tabitem Who tells Caleb that Joshua has the ring? -\color{tp2}{\textbf{Livvie}} \\ 
& \tabitem Who realizes Frank is a vampire? -\color{tp2}{\textbf{Karen}} \\
& \tabitem What does Livvie tell Caleb Joshua has? -\color{tp2}{\textbf{the ring}} \\
& \tabitem Who does Karen realize Frank is? -
\color{tp2}{\textbf{vampire}} \\ \hline \hline
\textbf{CS (text-only)} & Caleb and Rafe discuss how to get close to Joshua and Livvie. Lucy tries to convince Victor that Joshua is an evil vampire who should not be allowed to have his soul. Lucy tells Victor that she can t lose him and wants him to accept her offer to turn him back into a vampire. Joshua tells the people of Port Charles that he will do whatever it takes to breathe new life into this wonderful old place. \\ 
\multirow{6}{5em}{\textbf{QA pairs}} & \tabitem Who tells Lucy that he enjoys the power Joshua has given him? -\color{tp2}{\textbf{Victor}} \\
& \tabitem Who does Karen realize is a vampire? -\textcolor{red}{\textbf{Joshua}} \\
& \tabitem Who pleads with Victor to fight Joshua? -\color{tp2}{\textbf{Lucy}} \\
& \tabitem Who tells Caleb that Joshua has the ring? -\textcolor{red}{\textbf{Rafe}} \\ 
& \tabitem Who realizes Frank is a vampire? -\textcolor{red}{\textbf{Victor}} \\
& \tabitem What does Livvie tell Caleb Joshua has? -\textcolor{red}{\textbf{soul}} \\
& \tabitem Who does Karen realize Frank is? -
\color{tp2}{\textbf{vampire}} \\ \hline \hline
\textbf{TPs (text-only)} & Caleb and Livvie are shocked to find out that the ring has been taken away from them by Joshua. They are unable to get the ring back, but they are determined to find a way to get it back. Lucy tells Victor that Joshua is a liar and that he should not be allowed to have an important position in Port Charles. Victor tells Lucy that he will not give up on her, but she tells him that she will not go to the hospital because she has to be here for Victor. Lucy and Victor agree that they need to stay at the hospital for the sake of Victor, but Lucy is not willing to give up her job as Victor s guardian angel. Rafe tells Alison that if she turns back, she is going to be under Joshua s control, and he doesn t think she will be too happy about taking orders from someone else. He tells her that he would rather slit her own wrists than have Joshua touch her. \\
\multirow{7}{5em}{\textbf{QA pairs}} & \tabitem Who tells Lucy that he enjoys the power Joshua has given him? -\color{tp2}{\textbf{Victor}} \\
& \tabitem Who does Karen realize is a vampire? -\textcolor{red}{\textbf{Joshua}} \\
& \tabitem Who pleads with Victor to fight Joshua? -\color{tp2}{\textbf{Lucy}} \\
& \tabitem Who tells Caleb that Joshua has the ring? -\textcolor{red}{\textbf{Lucy}} \\ 
& \tabitem Who realizes Frank is a vampire? -\textcolor{red}{\textbf{Lucy}} \\
& \tabitem What does Livvie tell Caleb Joshua has? -\color{tp2}{\textbf{the ring}} \\
& \tabitem Who does Karen realize Frank is? -
\color{tp2}{\textbf{vampire}} \\ \hline \hline
\textbf{Pseudo-oracle} & Caleb and Livvie find out that the ring that protects them from Joshua is gone, and they have no way of getting it back. Lucy and Ian try to convince Victor to turn Joshua into a vampire, but he tells them that he is not a monster and that he wants to be a good man. Lucy tells Victor that Joshua is evil, but Victor tells her that she is his friend and he will not let her get to him. Victor tells Lucy that he will do whatever it takes to breathe new life into Port Charles. Ricky tells Jamal that he does not want to be his guardian angel anymore, but Jamal tells him to go and hang with the bloodsuckers. Jack tells Jack that the only way he is going to win is to play both sides of the fence. Frank tells Karen that there are benefits to his new role as a vampire and he is willing to do it for the good of the town of Port Charles and his family.  \\
\multirow{7}{5em}{\textbf{QA pairs}} & \tabitem Who tells Lucy that he enjoys the power Joshua has given him? -\color{tp2}{\textbf{Victor}} \\
& \tabitem Who does Karen realize is a vampire? -\color{tp2}{\textbf{Frank}} \\
& \tabitem Who pleads with Victor to fight Joshua? -\color{tp2}{\textbf{Lucy}} \\
& \tabitem Who tells Caleb that Joshua has the ring? -\color{tp2}{\textbf{Livvie}} \\ 
& \tabitem Who realizes Frank is a vampire? -\color{tp2}{\textbf{Karen}} \\
& \tabitem What does Livvie tell Caleb Joshua has? -\color{tp2}{\textbf{the ring}} \\
& \tabitem Who does Karen realize Frank is? - \color{tp2}{\textbf{vampire}} \\
\hline
\end{tabular}
\caption{Gold summary with automatically generated QA pairs (top) and model
  summaries with different content selection methods. Questions which
  the automatic summary answers correctly are highlighted in green
  (wrong answers shown in red).  All model variants consider the
  textual modality only (i.e.,~BART with adapter tuning).}
\label{tab:examples_summaries_content_selection}
\end{table*}

\begin{table*}[t]
\small
\centering
\begin{tabular}{L{5em}L{40em}}
\hline
\textbf{Model} & \multicolumn{1}{c}{\textbf{Summary}} \\ 
\hline
\textbf{Gold} & Caleb is upset when Livvie tells him that Joshua has the ring.  Joshua attempts to sway Ian to the dark side, but Ian vows he will
 continue to fight Joshua and the other vampires.  Rafe tells Caleb the only way he can defeat Joshua now is to remain human and Livvie reluctantly agrees.  Lucy pleads with Victor to fight Joshua, however, it s too late, as Victor tells her he enjoys the power Joshua has given him. Karen realizes Frank is a vampire. \\
\multirow{7}{5em}{\textbf{QA pairs}} & \tabitem Who tells Lucy that he enjoys the power Joshua has given him? -\color{tp2}{\textbf{Victor}} \\
& \tabitem Who does Karen realize is a vampire? -\color{tp2}{\textbf{Frank}} \\
& \tabitem Who pleads with Victor to fight Joshua? -\color{tp2}{\textbf{Lucy}} \\
& \tabitem Who tells Caleb that Joshua has the ring? -\color{tp2}{\textbf{Livvie}} \\ 
& \tabitem Who realizes Frank is a vampire? -\color{tp2}{\textbf{Karen}} \\
& \tabitem What does Livvie tell Caleb Joshua has? -\color{tp2}{\textbf{the ring}} \\
& \tabitem Who does Karen realize Frank is? - \color{tp2}{\textbf{vampire}} \\ \hdashline
\textbf{Text-only (TPs)} & Caleb and Livvie are shocked to find out that the ring has been taken away from them by Joshua. They are unable to get the ring back, but they are determined to find a way to get it back. Lucy tells Victor that Joshua is a liar and that he should not be allowed to have an important position in Port Charles. Victor tells Lucy that he will not give up on her, but she tells him that she will not go to the hospital because she has to be here for Victor. Lucy and Victor agree that they need to stay at the hospital for the sake of Victor, but Lucy is not willing to give up her job as Victor s guardian angel. Rafe tells Alison that if she turns back, she is going to be under Joshua s control, and he doesn t think she will be too happy about taking orders from someone else. He tells her that he would rather slit her own wrists than have Joshua touch her. \\ 
\multirow{7}{5em}{\textbf{QA pairs}} & \tabitem Who tells Lucy that he enjoys the power Joshua has given him? -\color{tp2}{\textbf{Victor}} \\
& \tabitem Who does Karen realize is a vampire? -\textcolor{red}{\textbf{Joshua}} \\
& \tabitem Who pleads with Victor to fight Joshua? -\color{tp2}{\textbf{Lucy}} \\
& \tabitem Who tells Caleb that Joshua has the ring? -\textcolor{red}{\textbf{Lucy}} \\ 
& \tabitem Who realizes Frank is a vampire? -\textcolor{red}{\textbf{Lucy}} \\
& \tabitem What does Livvie tell Caleb Joshua has? -\color{tp2}{\textbf{the ring}} \\
& \tabitem Who does Karen realize Frank is? -
\color{tp2}{\textbf{vampire}} \\ \hline \hline
\textbf{Hierarchical3D (TPs)} & Caleb and Livvie are shocked when they find out that their protection against Joshua is gone. Victor and Lucy argue about Victor's role in Port Charles. Lucy tells Victor that Joshua is evil and that he should not be allowed to have an important position with 
the vampires. Victor tells Lucy that he still has so much to contribute and maybe this is his chance to have people listen to him 
again. Lucy is upset that Victor wants to give Joshua an important role in the town. Lucy and Victor are at the hospital and Lucy tells him that she is going to be there for Victor, but he tells her to stay away from him. Frank tells Karen that he has grown a pair of fangs. Karen is shocked to hear that Frank is a vampire. \\ 
\multirow{7}{5em}{\textbf{QA pairs}} & \tabitem Who tells Lucy that he enjoys the power Joshua has given him? -\color{tp2}{\textbf{Victor}} \\
& \tabitem Who does Karen realize is a vampire? -\color{tp2}{\textbf{Frank}} \\
& \tabitem Who pleads with Victor to fight Joshua? -\color{tp2}{\textbf{Lucy}} \\
& \tabitem Who tells Caleb that Joshua has the ring? -\textcolor{red}{\textbf{Lucy}} \\ 
& \tabitem Who realizes Frank is a vampire? -\color{tp2}{\textbf{Karen}} \\
& \tabitem What does Livvie tell Caleb Joshua has? -\textcolor{red}{\textbf{their protection against Joshua}} \\
& \tabitem Who does Karen realize Frank is? - \color{tp2}{\textbf{vampire}} \\ \hline \hline
\textbf{LED} & At the end of the night, Caleb and Livvie s wedding ring is revealed to Rafe and Caleb s ring. Caleb tells Rafe that the ring is a vampire slayer. Rafe tells Caleb that he s going to be a vampire again.    Lucy tells Victor that Victor is going to take over the town of Port Charles.  Victor tells Lucy that he wants to talk to Lucy.  Lucy asks Victor to join her in the new venture, but Victor tells her that he is not going to do it.  Caleb tells Olivia that he has been drugged by Kevin, and he s been turned into a vampire.  Olivia tells him that she wants to be part of the new club, but Caleb tells her to stay away from him.  Joshua tells Ian that he will not be able to get Victor away from Victor.  Ian tells Joshua that Joshua is not one of the vampire slayers, but he is the one of them. \\ 
\multirow{7}{5em}{\textbf{QA pairs}} & \tabitem Who tells Lucy that he enjoys the power Joshua has given him? -\color{tp2}{\textbf{Victor}} \\
& \tabitem Who does Karen realize is a vampire? -\textcolor{red}{\textbf{Caleb}} \\
& \tabitem Who pleads with Victor to fight Joshua? -\textcolor{red}{\textbf{Ian}} \\
& \tabitem Who tells Caleb that Joshua has the ring? -\textcolor{red}{\textbf{Ian}} \\ 
& \tabitem Who realizes Frank is a vampire? -\textcolor{red}{\textbf{Rafe}} \\
& \tabitem What does Livvie tell Caleb Joshua has? -\color{tp2}{\textbf{wedding ring}} \\
& \tabitem Who does Karen realize Frank is? - \textcolor{red}{\textbf{slayer}} \\
\hline
\end{tabular}
\caption{Gold summary with automatically generated QA pairs (top) and model
  summaries. Questions which
  the automatic summary answers correctly are highlighted in green
  (wrong answers shown in red).  We compare our approach (i.e.,~Hierarchical3D) with state-of-the-art textual summarizers (i.e.,~LED).}
\label{tab:examples_summaries_comparison_systems}
\end{table*}

\begin{table*}[t]
\small
\centering
\begin{tabular}{L{5em}L{40em}}
\hline
\textbf{Model} & \multicolumn{1}{c}{\textbf{Summary}} \\ 
\hline
\textbf{Gold} & Joshua tells Elizabeth he wants to turn Allison and demands she help ease Allison into her new life as his wife.  Elizabeth tells 
Joshua she will kill him before she allows him to hurt Allison.  Livvie is able to fend off her need to feed while she and Caleb make love.  Frank searches for Allison.  When Frank attempts to kidnap Allison from Rafe, he discovers that it really is Lucy and I
an in disguise.  Allison and Rafe reappear in Caleb s cave. \\
\multirow{7}{5em}{\textbf{QA pairs}} & \tabitem Who does Frank try to kidnap Allison from? -\color{tp2}{\textbf{Rafe}} \\
& \tabitem Who does Frank try to kidnap? -\color{tp2}{\textbf{Allison}} \\ 
& \tabitem Who tries to kidnap Allison? -\color{tp2}{\textbf{Frank}} \\
& \tabitem Who can fend off her need to feed while she and Caleb make love? -\color{tp2}{\textbf{Livvie}} \\
& \tabitem Who tells Joshua she will kill him before she allows him to hurt Allison? -\color{tp2}{\textbf{Elizabeth}} \\
& \tabitem Who tells Elizabeth he wants to turn Allison into his wife? -\color{tp2}{\textbf{Joshua}} \\ 
& \tabitem What is Allison s new life? -\color{tp2}{\textbf{wife}} \\ \hline \hline
\textbf{CS (text-only)} & Rafe tells Alison that he will never let Joshua take her for his bride, but she tells him that she has no choice in the matter. Elizabeth tells Joshua that she will not stand by and allow him to take her daughter. Joshua tells Elizabeth that he is going to eas
e Alison into her new lifestyle as his wife. Elizabeth says that she is not going to let her daughter suffer the kind of nightmare
 that she lived. She will kill Joshua before he is even that close to turning her. Alison tells Rafe that she thinks this is a little extreme, that is all. Rafe says he will not let Joshua get to her. He promises to keep her away from Joshua and all his goons.
 Caleb tells Livvie that she doesn t need to feed. He tells her that he can t make love to her the way she wants to. She tells him
 she can t turn him back, but he tells her he can. He says that he loves her and that he wants to make her his bride. \\ 
\multirow{7}{5em}{\textbf{QA pairs}} & \tabitem Who does Frank try to kidnap Allison from? -\textcolor{red}{\textbf{Joshua}} \\
& \tabitem Who does Frank try to kidnap? -\textcolor{red}{\textbf{Joshua}} \\ 
& \tabitem Who tries to kidnap Allison? -\textcolor{red}{\textbf{Rafe}} \\
& \tabitem Who can fend off her need to feed while she and Caleb make love? -\color{tp2}{\textbf{Livvie}} \\
& \tabitem Who tells Joshua she will kill him before she allows him to hurt Allison? -\color{tp2}{\textbf{Elizabeth}} \\
& \tabitem Who tells Elizabeth he wants to turn Allison into his wife? -\color{tp2}{\textbf{Joshua}} \\ 
& \tabitem What is Allison s new life? -\color{tp2}{\textbf{wife}} \\ \hline \hline
\textbf{TPs (text-only)} & Livvie tells Caleb that she can t be with him, knowing what his bite might do to him. Joshua tells Elizabeth that he is going to steal Alison s slayer s intended and that will make him more respectable. He tells her that she is not going to do it, and that he 
knows that both roads lead Alison right back to Joshua. Elizabeth tells Joshua that she has no idea how much she would love to do 
that, but they both know that they are too busy figuring out which road to take. Rafe and Alison find themselves in Caleb s old hangout. Alison is shocked to see that Rafe is still there. She tells Rafe that she will feel better when all this is over with. She asks Rafe if he has his back and he says that he does, and he tells her he has her back. Lucy and Ian find out that they have succeeded in their mission to kidnap Alison. Ian and Frank tell Lucy that they will have to tell their boss that their mission didn t
 work. Frank tells Ian that he will pay for this, and they are both going to pay for it. Ian tells Frank that he has to tell his big, powerful boss that his mission failed. \\ 
\multirow{7}{5em}{\textbf{QA pairs}} & \tabitem Who does Frank try to kidnap Allison from? -\textcolor{red}{\textbf{Lucy}} \\
& \tabitem Who does Frank try to kidnap? -\color{tp2}{\textbf{Alison}} \\ 
& \tabitem Who tries to kidnap Allison? -\textcolor{red}{\textbf{Lucy}} \\
& \tabitem Who can fend off her need to feed while she and Caleb make love? -\textcolor{red}{\textbf{Rafe}} \\
& \tabitem Who tells Joshua she will kill him before she allows him to hurt Allison? -\color{tp2}{\textbf{Elizabeth}} \\
& \tabitem Who tells Elizabeth he wants to turn Allison into his wife? -\color{tp2}{\textbf{Joshua}} \\ 
& \tabitem What is Allison s new life? -\textcolor{red}{\textbf{Caleb}} \\ \hline \hline
\textbf{Pseudo-oracle} & Caleb tells Livvie that he loves her and wants to make love to her, but he can t because his fangs are poisoned and he can s not turn her back. He tells her to take the risk, but she is afraid to do it. She tells him that she loves him and that she wants to be with him, but that she can t do it because of the risk of his bite. He says that he will not let her go back to him. She says that she will never stop loving him. He promises her that he won t and they make love again. Joshua tells Elizabeth that he wants to take Alison as his bride. Elizabeth tells Joshua that she is not going to let him take her. She wants to stop him from taking Alison. Rafe tells Alison that he has a vision of a city of tortured souls, a master vampire race, and his vampire bride. He will never let that happen. He can t let Joshua get to her. He is going to keep her away from Joshua and all his goons. He wants her to be his bride and she will be his wife. She will never love him, Rafe says.  \\
\multirow{7}{5em}{\textbf{QA pairs}} & \tabitem Who does Frank try to kidnap Allison from? -\textcolor{red}{\textbf{Joshua}} \\
& \tabitem Who does Frank try to kidnap? -\color{tp2}{\textbf{Allison}} \\ 
& \tabitem Who tries to kidnap Allison? -\textcolor{red}{\textbf{Joshua}} \\
& \tabitem Who can fend off her need to feed while she and Caleb make love? -\color{tp2}{\textbf{Livvie}} \\
& \tabitem Who tells Joshua she will kill him before she allows him to hurt Allison? -\color{tp2}{\textbf{Elizabeth}} \\
& \tabitem Who tells Elizabeth he wants to turn Allison into his wife? -\color{tp2}{\textbf{Joshua}} \\ 
& \tabitem What is Allison s new life? -\color{tp2}{\textbf{vampire bride}} \\
\hline
\end{tabular}
\caption{
  Gold summary with automatically generated QA pairs (top) and model
  summaries with different content selection methods. Questions which
  the automatic summary answers correctly are highlighted in green
  (wrong answers shown in red).   All model variants consider the textual modality (i.e.,~BART with adapter tuning).}
\label{tab:examples_summaries_content_selection_2}
\end{table*}

\begin{table*}[t]
\small
\centering
\begin{tabular}{L{5em}L{40em}}
\hline
\textbf{Model} & \textbf{Summary} \\ 
\hline
\textbf{Gold} & Joshua tells Elizabeth he wants to turn Allison and demands she help ease Allison into her new life as his wife.  Elizabeth tells 
Joshua she will kill him before she allows him to hurt Allison.  Livvie is able to fend off her need to feed while she and Caleb make love.  Frank searches for Allison.  When Frank attempts to kidnap Allison from Rafe, he discovers that it really is Lucy and I
an in disguise.  Allison and Rafe reappear in Caleb s cave. \\ 
\multirow{7}{5em}{\textbf{QA pairs}} & \tabitem Who does Frank try to kidnap Allison from? -\color{tp2}{\textbf{Rafe}} \\
& \tabitem Who does Frank try to kidnap? -\color{tp2}{\textbf{Allison}} \\ 
& \tabitem Who tries to kidnap Allison? -\color{tp2}{\textbf{Frank}} \\
& \tabitem Who can fend off her need to feed while she and Caleb make love? -\color{tp2}{\textbf{Livvie}} \\
& \tabitem Who tells Joshua she will kill him before she allows him to hurt Allison? -\color{tp2}{\textbf{Elizabeth}} \\
& \tabitem Who tells Elizabeth he wants to turn Allison into his wife? -\color{tp2}{\textbf{Joshua}} \\ 
& \tabitem What is Allison s new life? -\color{tp2}{\textbf{wife}} \\ \hline \hline
\textbf{Text-only (CS)} & Rafe tells Alison that he will never let Joshua take her for his bride, but she tells him that she has no choice in the matter. Elizabeth tells Joshua that she will not stand by and allow him to take her daughter. Joshua tells Elizabeth that he is going to eas
e Alison into her new lifestyle as his wife. Elizabeth says that she is not going to let her daughter suffer the kind of nightmare
 that she lived. She will kill Joshua before he is even that close to turning her. Alison tells Rafe that she thinks this is a little extreme, that is all. Rafe says he will not let Joshua get to her. He promises to keep her away from Joshua and all his goons.
 Caleb tells Livvie that she doesn t need to feed. He tells her that he can t make love to her the way she wants to. She tells him
 she can t turn him back, but he tells her he can. He says that he loves her and that he wants to make her his bride. \\ 
\multirow{7}{5em}{\textbf{QA pairs}} & \tabitem Who does Frank try to kidnap Allison from? -\textcolor{red}{\textbf{Joshua}} \\
& \tabitem Who does Frank try to kidnap? -\textcolor{red}{\textbf{Joshua}} \\ 
& \tabitem Who tries to kidnap Allison? -\textcolor{red}{\textbf{Rafe}} \\
& \tabitem Who can fend off her need to feed while she and Caleb make love? -\color{tp2}{\textbf{Livvie}} \\
& \tabitem Who tells Joshua she will kill him before she allows him to hurt Allison? -\color{tp2}{\textbf{Elizabeth}} \\
& \tabitem Who tells Elizabeth he wants to turn Allison into his wife? -\color{tp2}{\textbf{Joshua}} \\ 
& \tabitem What is Allison s new life? -\color{tp2}{\textbf{wife}} \\ \hline \hline
\textbf{Hierarchical3D (CS)} & Livvie tries to convince Caleb to let her take the risk of biting him, but she is afraid that she won t be able to do it. Joshua tells Elizabeth that he wants Alison to be his bride. Elizabeth is shocked when she finds out that Joshua wants to take Alison away
 from Rafe. Elizabeth tells Joshua that she will find a way to stop him from taking Alison. Rafe tells Alison that he has a vision
 of a city of tortured souls, a master vampire race, and his vampire bride. He tells her that he can make a perfect bride for her.
 Alison tells Rafe that she doesn t want to leave her family, but Rafe assures her that she is not going to leave them. Frank tells Ian that he is going to have to tell his boss that his mission didn t work. \\ 
\multirow{7}{5em}{\textbf{QA pairs}} & \tabitem Who does Frank try to kidnap Allison from? -\color{tp2}{\textbf{Rafe}} \\
& \tabitem Who does Frank try to kidnap? -\color{tp2}{\textbf{Allison}} \\ 
& \tabitem Who tries to kidnap Allison? -\textcolor{red}{\textbf{Rafe}} \\
& \tabitem Who can fend off her need to feed while she and Caleb make love? -\color{tp2}{\textbf{Livvie}} \\
& \tabitem Who tells Joshua she will kill him before she allows him to hurt Allison? -\color{tp2}{\textbf{Elizabeth}} \\
& \tabitem Who tells Elizabeth he wants to turn Allison into his wife? -\color{tp2}{\textbf{Joshua}} \\ 
& \tabitem What is Allison s new life? -\color{tp2}{\textbf{vampire bride}} \\ \hline \hline
\textbf{LED} & Caleb and Livvie try to convince Caleb that they can be together again, but Caleb refuses to believe that he can t make love to Livvie. Rafe and Alison agree that they will be together, but Rafe tells Alison that he will never be able to make love with her again. Elizabeth tells Joshua that she wants him to turn Alison into a vampire bride. Joshua tells Elizabeth that he wants her to be his wife, but Elizabeth tells him that she is not going to do that. Joshua says that he is going to make Alison a vampire, and he will be the one true love of his life. Elizabeth says that she will never let Alison suffer the kind of nightmare that she lived, and she will make her life as a vampire. Joshua and Elizabeth argue about how much she wants to be a vampire and how much he wants to help her. Elizabeth asks Joshua if he s going to help Alison, but he says he will not. \\
\multirow{7}{5em}{\textbf{QA pairs}} & \tabitem Who does Frank try to kidnap Allison from? -\textcolor{red}{\textbf{Caleb}} \\
& \tabitem Who does Frank try to kidnap? -\textcolor{red}{\textbf{Caleb}} \\ 
& \tabitem Who tries to kidnap Allison? -\textcolor{red}{\textbf{Rafe}} \\
& \tabitem Who can fend off her need to feed while she and Caleb make love? -\color{tp2}{\textbf{Livvie}} \\
& \tabitem Who tells Joshua she will kill him before she allows him to hurt Allison? -\color{tp2}{\textbf{Elizabeth}} \\
& \tabitem Who tells Elizabeth he wants to turn Allison into his wife? -\color{tp2}{\textbf{Joshua}} \\ 
& \tabitem What is Allison s new life? -\textcolor{red}{\textbf{vampire}} \\
\hline
\end{tabular}
\caption{Gold summary with automatically generated QA pairs (top) and model
  summaries. Questions which
  the automatic summary answers correctly are highlighted in green
  (wrong answers shown in red).  We compare our approach (i.e.,~Hierarchical3D) with state-of-the-art textual summarizers (i.e.,~LED).}
\label{tab:examples_summaries_comparison_systems_2}
\end{table*}

\subsection{Examples of Generated Summaries} \label{sec:appendix_examples}

In this section we provide examples of generated summaries based on
different automatic systems. Moreover, we provide examples of
questions and answers used for the automatic QA evaluation described
in Section~\ref{sec:evaluation_metrics}.

Table~\ref{tab:examples_QA} shows examples of the automatically
generated question-answer pairs given gold standard summaries. We
provide examples of QA pairs for named entities (first 4 rows of the
table) and nouns (remaining 6 rows of the table). We observe that most
QA pairs are reasonable and correspond to information given in
human-written summaries (first column of the table). However, there
are cases where the QA pairs do not provide reasonable questions. Such
an example is illustrated in the last row of
Table~\ref{tab:examples_QA}, where the question is generated given the
summary segment ``Jonathan and Lizzie find out their baby has a
medical condition, and make a run for it'':
\begin{itemize}
    \item[] \hspace{-1em} \textbf{Q}: ``What do Lizzie and Jonathan do when they learn their baby has a medical condition?''
    \item[] \hspace{-1em} \textbf{A}: ``run''
\end{itemize}
This QA pair does not correspond to a reasonable fact of the
episode. This shows that although it is useful to filter the
questions, there are still imperfections with the automatic generation
of QA pairs, especially when considering nouns.

Next, we give examples of the generated summaries for the TV show
''Port Charles'' in
Tables~\mbox{\ref{tab:examples_summaries_content_selection}--\ref{tab:examples_summaries_comparison_systems_2}}. In
every case, we present the gold or generated summary alongside the QA
pairs used for evaluation. First, we compare different content
selection methods (i.e.,~supervised content selection (CS), TP
identification (TPs), and pseudo-oracle) for a text-only summarizer
based on BART with adapter tuning. We present two examples in
Tables~\ref{tab:examples_summaries_content_selection}
and~\ref{tab:examples_summaries_content_selection_2} (we also show
gold summaries for each episode). In both cases, we observe that the
pseudo-oracle selection provides summaries of better quality, with
fewer errors in the questions answered (i.e.,~errors are illustrated
with {\color{red}red}). Moreover, when comparing content selection
(CS) with TP identification (TPs), we find that these two approaches
provide similar results, as suggested by our main experimental results
(Table~\ref{tab:automatic_qa_test_set}). Specifically, in
Table~\ref{tab:examples_summaries_content_selection}, TP
identification seems to provide the most informative summary, whereas
in Table~\ref{tab:examples_summaries_content_selection_2} supervised
content selection is the best option.

Secondly, we compare our approach that considers multimodal
information (Hierarchical3D) against text-only BART with equivalent
content selection, and LED which considers only text and uses an
efficient self-attention mechanism for processing the entire input. We
present two examples for the same episodes as above in
Tables~\ref{tab:examples_summaries_comparison_systems}
and~\ref{tab:examples_summaries_comparison_systems_2}. We empirically
validate that the quality of the generated summaries is improved by
adding the multimodal information (both when using supervised content
selection and TP identification). Our approach leads to summaries that
 answer correctly a larger percentage of automatic
questions (i.e.,~correct answers are illustrated with
{\color{tp2}green}) outperforming LED, which is fully fine-tuned and
memory-heavy. Interestingly, LED summaries cannot answer a
large proportion of the given questions, suggesting that such methods
may not be suitable for the task and small dataset size.

\end{document}